%% file: main.tex
\PassOptionsToPackage{dvipsnames}{xcolor}
\documentclass[nonacm]{acmart}

\AtBeginDocument{%
  }

\setcopyright{acmlicensed}
\copyrightyear{2018}
\acmYear{2018}
\acmDOI{XXXXXXX.XXXXXXX}

\acmConference[Conference acronym 'XX]{Make sure to enter the correct
  conference title from your rights confirmation emai}{June 03--05,
  2018}{Woodstock, NY}

\acmISBN{978-1-4503-XXXX-X/18/06}

\usepackage{natbib}
\setcitestyle{numbers}

\usepackage{csquotes}
\usepackage{array}
\usepackage{amsmath}
\usepackage{amsfonts}
\usepackage{amssymb}
\usepackage{caption}
\usepackage{comment}
\usepackage[inline]{enumitem}
\usepackage{derivative}
\usepackage{subcaption}
\usepackage{algorithmic}
\usepackage{graphicx}
\usepackage{textcomp}
\usepackage{multirow}  
\usepackage{array}     
\usepackage{longtable}
\usepackage{bbding} 
\usepackage{adjustbox}

\def\equationautorefname~#1\null{Eq.~(#1)\null}

\input{macros}

\newcommand\subwidth{0.08\textwidth}
\newcolumntype{P}[1]{>{\centering\arraybackslash}p{#1}}
\newcommand{\yes}{\textcolor{ForestGreen}{\CheckmarkBold}}
\newcommand{\no}{\textcolor{BrickRed}{\XSolidBrush}}

\begin{document}

\title{Deep Model Merging: The Sister of Neural Network Interpretability - A Survey}


\input{authors}



\begin{abstract}
We survey the model merging literature through the lens of loss landscape geometry to connect observations from empirical studies on model merging and loss landscape analysis to phenomena that govern neural network training and the emergence of their inner representations. We distill repeated empirical observations from the literature in these fields into descriptions of four major characteristics of loss landscape geometry: mode convexity, determinism, directedness, and connectivity. We argue that insights into the structure of learned representations from model merging have applications to model interpretability and robustness, subsequently we propose promising new research directions at the intersection of these fields.
\end{abstract}

\ccsdesc[500]{Computing methodologies~Machine learning algorithms}
\ccsdesc[300]{Computing methodologies~Learning paradigms}
\ccsdesc[300]{Computing methodologies~Learning settings}
\ccsdesc[300]{Computing methodologies~Artificial intelligence}

\keywords{
    model merging, representation learning, deep learning, federated learning, interpolation, neural networks, optimization, stochastic gradient descent, loss landscape
}

\maketitle

\input{sections/1_intro}

\input{sections/2_background}

\input{sections/3_survey}
\input{sections/4_insights}

\input{sections/5_discussion}
\input{sections/6_future_work}

\input{sections/algorithm}
\begin{acks}
    This material is based on work supported by the U.S. Department of Energy, Office of Science, Office of Advanced Scientific Computing Research, Department of Energy Computational Science Graduate Fellowship under Award Number DE-SC0023112 and by the U.S. Department of Energy, Office of Science, Advanced Scientific Computing Research, under Contract DE-AC02-06CH11357.
\end{acks}


\bibliographystyle{ACM-Reference-Format}
\bibliography{refs/zotero}



\end{document}

%% file: authors.tex
\author{Arham Khan}
\orcid{1234-5678-9012}
\affiliation{%
  \institution{University of Chicago}
  \city{Chicago}
  \state{Illinois}
  \country{USA}
}
\email{arham@uchicago.edu}

\author{Todd Nief}
\affiliation{%
  \institution{University of Chicago}
  \city{Chicago}
  \state{Illinois}
  \country{USA}
}

\author{Nathaniel Hudson}
\affiliation{%
  \institution{University of Chicago}
  \city{Chicago}
  \state{Illinois}
  \country{USA}
}

\author{Mansi Sakarvadia}
\affiliation{%
  \institution{University of Chicago}
  \city{Chicago}
  \state{Illinois}
  \country{USA}
}

\author{Daniel Grzenda}
\affiliation{%
  \institution{University of Chicago}
  \city{Chicago}
  \state{Illinois}
  \country{USA}
}

\author{Aswathy Ajith}
\affiliation{%
  \institution{University of Chicago}
  \city{Chicago}
  \state{Illinois}
  \country{USA}
}

\author{Jordan Pettyjohn}
\affiliation{%
  \institution{Colorado School of Mines}
  \city{Golden}
  \state{Colorado}
  \country{USA}
}

\author{Kyle Chard}
\affiliation{%
  \institution{University of Chicago}
  \city{Chicago}
  \state{Illinois}
  \country{USA}
}

\author{Ian Foster}
\affiliation{%
  \institution{Argonne National Laboratory}
  \city{Lemont}
  \state{Illinois}
  \country{USA}
}

\renewcommand{\shortauthors}{Khan et al.}

%% file: sections/1_intro.tex
\section{Introduction}
\textit{Model merging} explores procedures for altering and combining the weights of neural network models to produce more performant and generalizable predictors. The model merging problem is as follows: Given the parameters of several \textit{source models} $\theta_1, \theta_2, \ldots, \theta_n$, we seek to develop a merging procedure $f(\theta_1, \theta_2, \ldots, \theta_n) : \Theta^n \mapsto \Theta$ to produce a new \textit{merged model} $\theta^\star$ that minimizes some loss $\mathcal{L}(\theta)$. That is, model merging methods produce a single model by combining the parameters of several previously trained models. 

Merging models is inherently challenging: we must account for symmetries in neural network architectures \cite{entezari2021role} (which lead to the existence of many behaviorally equivalent but mathematically distinct solutions to the same learning problem), understand how task-specific knowledge is stored in learnable parameters \cite{arous2023high, aghajanyan2020intrinsic}, and comprehend how changes in a model's training setup affect its final learned representations \cite{chaudhry2020continual}. One of the core challenges of model merging is that, barring certain exceptions, naively averaging model parameters leads to destructive interference---the tendency for merged parameters to overwrite or destroy information present in the original source models. As a result, model merging research explores techniques for weighting or aligning parameters in source models to avoid destructive interference when merging. This is often accomplished by distinguishing task-specific parameters from those that are generic or redundant. For example, when merging language models trained on two different languages, we might want to differentiate between parameters that encode language-specific knowledge and those that encapsulate higher-level semantic reasoning. Serendipitously, in the process of trying to prevent destructive interference, model merging algorithms yield procedures for mapping neural network components (e.g., layers, neurons, or weights) to task-specific behaviors. This ability to map learned parameters to semantic behaviors is one of the overarching goals of deep learning interpretability and robustness research. Therefore, studying model merging and developing more performant model merging techniques is tantamount to developing algorithms that can help explain neural network behavior by localizing distinct predictive behaviors in neural networks to subnetworks of their neurons.  
Indeed, many model merging techniques have shown promise in their ability to induce predictable and interpretable changes in model behavior \cite{wortsman2022model, ilharco2022editing, frankle2018lottery}. These can be considered parameter space analogs to popular interpretability techniques, such as activation steering or patching, that seek to identify and manipulate a model's activations (rather than weights) in ways that govern specific downstream behaviors \cite{turner2023activation, sakarvadia2023memory, vig2020causalmediationanalysisinterpreting}. Moreover, principles from model merging can yield insight into the susceptibility of models to adversaries \cite{shokri2017membership}, improve model robustness \cite{fradkin2022robustness, zhou2023going, xu2019understanding}, and decrease the risk of data leakage in the federated learning regime \cite{jin2022dataless}.

Recent advances in our understanding of model merging have shed light on each of these essential areas of neural network theory. Accordingly, we seek to present a comprehensive analysis and categorization of cutting-edge research from the model merging literature. In so doing, we distill insights on how neural network representations are structured and learned, and demonstrate how these insights could be applied to relate learned representations to model behavior (in the spirit of model interpretability) and enhance model robustness.

\subsection{Contributions}

Existing model merging surveys primarily focus on merging in specific regimes, such as federated learning, or focus only on application areas without connecting model merging to broader theoretical insights into deep learning.
For example, \citet{li2024federated} and \citet{yadav2024survey} focus their discussion of model merging on two particular application areas, namely federated learning and mixture-of-experts regimes, and include learning techniques, such as knowledge distillation, that rely principally on gradient descent rather than directly combining model weights. Meanwhile, \citet{yang2024model} taxonomize merging techniques according to when they are applied during the training pipeline, focusing on their relevance to application domains such as few-shot transfer, continual learning, and multimodal fusion. 
\citet{li2023deep} present a survey with a taxonomy similar to ours but focus on applying model merging to applications such as federated learning, knowledge distillation, and model fine-tuning.

In contrast, we present a taxonomy of model merging techniques organized into three major categories: ensembling, weight aggregation, and neuron alignment (see \autoref{fig:taxonomy}). We do not include techniques that require full training runs over a given dataset, such as knowledge distillation. Our work represents a philosophical departure from previous perspectives on model merging in that it focuses explicitly on extracting insights into how training proceeds under stochastic gradient descent (SGD), connecting the success of model merging techniques to the underlying structure of the loss landscape, and applying these ideas to model interpretability and robustness. We offer significant insights into the geometry of loss landscapes (see \autoref{sec:insights}) and its effects on model training beyond what previous model merging surveys have reported. In summary, we make the following contributions:


\begin{enumerate}
    \item We conduct a comprehensive review of the model merging literature and propose a 
    taxonomy of model merging techniques focused on their core algorithmic approach.
    \item We synthesize empirical data from many model merging studies to characterize phenomena that manifest in objective landscapes during training, illuminating how neural network representations are structured and learned.
    \item We draw meaningful connections between model merging, model interpretability, and model robustness, highlighting fruitful avenues for future research in these areas.
\end{enumerate}

\subsection{Survey Structure}

This work taxonomizes model merging techniques primarily based on their assumptions regarding the structure of learned representations. Therefore, we first divide the model merging literature into a set of three categories: ensembling, weight aggregation, and neuron alignment. Ensembling techniques combine only the output logits of each model, treating them purely as distributions and taking a Bayesian average over their parameters to obtain robust model predictions. Ensemble techniques are then further categorized into predictive techniques, which average model logits, and compositional techniques, which choose to compose model layers together during inference. Weight aggregation techniques are identified by their focus on computing linear combinations of model parameters to aggregate the parameters of each source model into a single, merged model. Weight aggregation techniques may do this in a simple manner, using a simple average of parameters, or using heuristic weightings, which may use various importance measures to determine which parameters ought to exert more influence in the final merged model. Finally, neuron alignment techniques aggregate weights but do so while addressing the potential issues brought on by permutation symmetries (see \autoref{bg_permutation_symmetry}) in neural network architectures. These techniques attempt to both weight and align model parameters to achieve effective merging and may do so by either searching for discrete permutation matrices or using real-valued alignment weightings between neurons to determine which parameters should be averaged together. Each of these sub-areas draws upon inspiration from the study of loss landscape geometry and provides empirical evidence explaining how model representations are structured as well as how those structures might vary under different training recipes. We aim to take these empirical insights and condense them into a set of principles that are useful for future research in deep learning interpretability and robustness.

To this end, the remainder of this paper is organized as follows: \autoref{sec:background} defines relevant terminology and introduces prerequisite concepts from the study of loss landscapes; Sections \ref{sec:ensembling}, \ref{sec:weight_averaging}, and \ref{permutation_alignment} survey research covering ensemble, weight aggregation, and neuron alignment techniques for model merging, respectively; \autoref{sec:insights} synthesizes the empirical observations from this survey into descriptions of loss landscape geometry to characterize commonly observed phenomena during model training; \autoref{sec:discussion} discusses the implications of our analysis on model interpretability and robustness; and \autoref{sec:future_work} proposes directions for future investigation in light of our analysis and concludes the survey. \autoref{tab:taxonomy} below lists and summarizes the techniques we survey in this work.

 \break

\input{graphics/tex/fig_taxonomy}

\input{sections/new_survey_table_reformatted}

%% file: graphics/tex/fig_taxonomy.tex
\begin{figure*}[t]
    \centering
    \includegraphics[width=\linewidth]{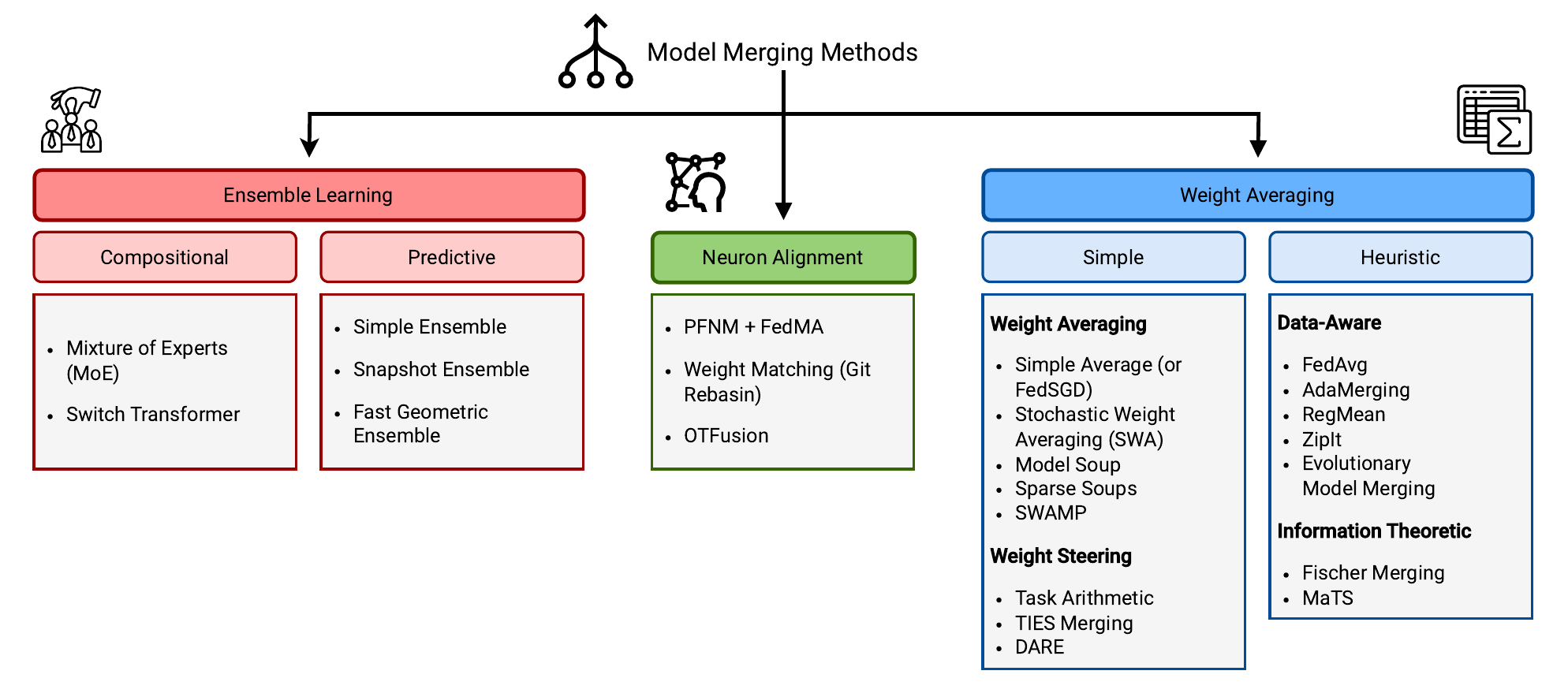}
    \caption{
        A taxonomy of model merging methods. \textbf{Ensemble} methods 
        do not manipulate model parameters, but rather combine prediction logits or model components in novel ways. 
        \textbf{Neuron Alignment} techniques attempt to resolve potential misalignment between model representations by using permutations of their units before computing some linear combination of source model parameters.
        \textbf{Weight Aggregation} methods compute a linear combination of model parameters.
    }
    \label{fig:taxonomy}
\end{figure*}

%% file: sections/new_survey_table_reformatted.tex
\footnotesize
\begin{longtable}{
    p{0.1\textwidth}
    p{0.1\textwidth}
    p{0.15\textwidth}
    p{0.14\textwidth}
    p{0.5\textwidth}
}
\caption{Table of Model Merging techniques covered in this work, categorized by their approach.} \\
\toprule
    Classification & Type & Name & Source & Summary \\
\midrule
\multirow{15}{\subwidth}{Ensembling}
  & \multirow{8}{\subwidth}{Predictive} 
  & Snapshot Ensembles
  & \citet{huang2017snapshot}
  & Cycles through learning rates during training to obtain several models in distinct minima, averages their logits during inference. \\
\cmidrule{3-5}
  &
  & Fast Geometric Ensembling
  & \citet{garipov2018loss}
  & Trains a single neural network traditionally for approximately 80\% of the time until convergence, then employs a dynamic learning rate schedule to obtain several models with low loss to eventually ensemble their logits. \\
\cmidrule{3-5}
  &
  & Simplices for Fast Ensembling
  & \citet{benton2021loss}
  & Trains several models to find a trivial simplicial complex of low loss solutions, then trains several \enquote{connecting} solutions such that we are left with a full simplex of low loss solutions for ensembling. \\
\cmidrule{2-5}
  & \multirow{6}{\subwidth}{Compositional}
  & Mixture-of-Experts (MoE)
  & \citet{shazeer2017outrageously}
  & MoE maintains a cohort of \enquote{experts}---specialized subnetworks of neurons---and learns a routing network which then selects the top-$k$ experts for any given input and outputs a linear combination of each of their predictions. \\
\cmidrule{3-5}
  &
  & Switch Transformers
  & \citet{fedus2022switch}
  &  Simplifies the MoE router by demonstrating that just a single expert per input example is sufficient for strong performance. Significantly reduces communication and computational costs enabling superior scaling of MoE models. \\
\midrule
\multirow{20}{\subwidth}{Weight Aggregation}
  & \multirow{14}{\subwidth}{Weight Averaging} 
  & FedSGD (Simple Average)
  & \citet{mcmahan2017communication} 
  & Computes a simple average over source model parameters. Requires models of the same architecture. \\
\cmidrule{3-5}
  & 
  & Stochastic Weighted Averaging 
  & \citet{izmailov2018averaging} 
  & Gathers checkpoints from various points in the training trajectory of a single model and computes a simple average over their parameters. \\
\cmidrule{3-5}
  & 
  & \multirow{2}{\subwidth}{Model Soup}
  & \citet{wortsman2022model} 
  & Takes the simple average of parameters from models of the same architecture fine-tuned with different hyperparameters. \\
\cmidrule{3-5}
  & 
  & Sparse Soup 
  & \citet{zimmer2023sparse} 
  & Constructs model soups that are more efficient for inference by applying a technique inspired by Iterative Magnitude Pruning~\cite{tanaka2020pruning} to produce sparse models. \\
\cmidrule{3-5}
  & 
  & SWAMP 
  & \citet{choi2023swamp} 
  & Iteratively applies Iterative Magnitude Pruning~\cite{tanaka2020pruning} and simple weight averaging between multiple fine-tuned models to produce a final sparse, merged model. \\
\cmidrule{2-5}
& \multirow{17}{\subwidth}{Weight Steering} 
  & \multirow{4}{\subwidth}{Task Arithmetic}
  & \citet{ilharco2022editing} 
  & Computes a \enquote{task vector}: the difference between the pre-trained and fine-tuned model weights for a particular task. Applying task vectors to pre-trained model weights via addition---and removing them via subtraction---is an intuitive way to alter model behavior and enhance or detract from performance on specific tasks. \\
\cmidrule{3-5}
  & 
  & \multirow{4}{\subwidth}{Model Breadcrumbs}
  & \citet{davari2023model} 
  & Takes task vectors from several fine-tuned models and limits destructive interference between task-specific parameters by masking outlier directions in each task vector. Suitable for applying Task Arithmetic~\cite{ilharco2022editing} to enhance performance in multiple tasks simultaneously. \\
\cmidrule{3-5}
  & 
  & \multirow{3}{\subwidth}{TIES-merging} 
  & \citet{yadav2024ties} 
  & Improves the performance and stability of Task Arithmetic~\cite{ilharco2022editing} when using multiple task vectors by identifying redundancies between task vectors, resolving sign conflicts between corresponding parameters in task vectors, and exclusively merging parameters that align with the final agreed-upon sign. \\
\cmidrule{3-5}
  & 
  & DARE
  & \citet{yu2024language} 
  & Builds upon Task Arithmetic~\cite{ilharco2022editing} by sparsifying task vectors, greatly reducing interference between merged parameters. DARE demonstrates that, in some cases, up to 90-99\% of fine-tuning parameters can be eliminated while preserving fine-tuning performance. This, in turn, can be exploited to efficiently create multi-task LLMs using model merging. \\
\cmidrule{2-5}
& \multirow{20}{\subwidth}{Data-Aware} 
  & FedAvg
  & \citet{sun2022decentralized} 
  & Computes a weighted average between source model parameters in a federated setting. Weights each source model's parameters by the proportion of the data in the federated system that is present at their node. Suitable for merging models whose data distributions are heterogeneous and non-IID. \\
\cmidrule{3-5}
  & 
  & AdaMerging
  & \citet{yang2023adamerging} 
  & Builds upon Task Arithmetic~\cite{ilharco2022editing}. Learns merging coefficients for task vectors adaptively by minimizing prediction entropy over unlabeled test data. Demonstrates strong multi-task performance, comparable with that of an ensemble. \\
\cmidrule{3-5}
  & 
  & \multirow{3}{\subwidth}{RegMean}
  & \citet{jin2022dataless} 
  & Merges weights at each layer by computing the least-squares fit to their activation maps. This ensures that the merged weight matrix produces activations similar to that of each source matrix when presented with data from that source model's training distribution.  \\
\cmidrule{3-5}
  & 
  & \multirow{4}{\subwidth}{ZipIt!}
  & \citet{stoica2023zipit} 
  & Identifies redundant parameters between models whose activations are highly correlated. This technique only merges redundant model parameters and preserves task-specific ones to prevent destructive interference when merging models. Demonstrates strong multi-task performance, even when models are trained independently.  \\
\cmidrule{3-5}
  & 
  & \multirow{3}{\subwidth}{Evolutionary Model Merging}
  & \citet{akiba2024evolutionary} 
  & Employs an evolutionary algorithm to learn whether to compose layers in sequence or merge them with a learned weighting. The final layer composition and weights for model merging are optimized to maximize training accuracy. \\
\cmidrule{2-5}
& \multirow{8}{\subwidth}{Information Theoretic} 
  & Fisher Weighted Averaging
  & \citet{matena2022merging} 
  & Computes a weighted average between source model parameters, with weights for each parameter defined by the Fisher information of that parameter. Applies a diagonal approximation of the Fisher Information matrix for computational tractability. \\
\cmidrule{3-5}
  & 
  & Merging by Matching Models in Task Parameter Subspaces (MaTS)
  & \citet{tam2023merging} 
  & Poses model merging as equivalent to solving a linear system of equations for the optimal merged parameter values. They design an enhanced Fisher Weighted Averaging procedure by using the Kronecker Factorization to obtain a more accurate block-diagonal approximation to the Fisher Information matrix. Computing the closed-form solution to this system is intractable in practice; therefore, they apply the method of Conjugate Gradients to find an approximate solution.  \\
\midrule
\multirow{29}{\subwidth}{Neuron Alignment}
& \multirow{20}{\subwidth}{Hard Alignment}  
  & Weight Matching 
  & \citet{ainsworth2022git}
  & Aligns corresponding layers of neurons between models of the same architecture by maximizing the similarity between their weight matrices (i.e., maximizing the Frobenius inner product of both weight matrices). They pose this problem as an NP-hard SOBLAP assignment problem and approximate a greedy solution using coordinate descent. Once weights are aligned, they are simply averaged. \\
\cmidrule{3-5}
  & 
  & Optimizing Mode Connectivity via Neuron Alignment
  & \citet{tatro2020optimizing} 
  & Presents a heuristic, ``neuron alignment'', which measures similarity between the distribution of activations between two weight matrices. They relate finding a permutation matrix which optimizes this metric to the assignment problem~\cite{burkard1999linear} and optimal transport. Weights are interpolated between once aligned.  \\
\cmidrule{3-5}
  & 
  & Probabilistic Federated Neural Matching (PFNM)
  & \citet{yurochkin2019bayesian} 
  & Aligns the parameters of edge models to a global model in a federated setting by using a non-parametric Bayesian model to determine the optimal assignment of neurons to one another for alignment and merging into a global model.  \\
\cmidrule{3-5}
  & 
  & Federated Learning with Matched Averaging (FedMA)
  & \citet{wang2020federated} 
  & Poses neuron alignment as an extended bipartite matching problem and applies the iterative Hungarian matching algorithm to solve the corresponding assignment problem, aligning neurons before merging.  \\
\cmidrule{2-5}
& \multirow{8}{\subwidth}{Soft Alignment}  
  & Optimal Transport Fusion (OTFusion) 
  & \citet{singh2020model}
  & Generalizes prior approaches which pose neuron alignment as a discrete assignment problem by modeling neuron alignment using the optimal transport framework. Use established techniques to solve the optimal transport problem to produce \enquote{soft} alignment maps between neurons, which can be used to align and combine corresponding layers between source models. \\
\cmidrule{3-5}
  & 
  & Transformer Fusion with Optimal Transport
  & \citet{imfeld2023transformer} 
  & Extends OTFusion~\cite{singh2020model} to handle modern Transformer architectures, also proposes a technique to align and merge models of heterogeneous architectures.  \\
\bottomrule
\label{tab:taxonomy}
\end{longtable}
\normalsize

%% file: sections/2_background.tex
\section{Background}
\label{sec:background}

Model merging is intimately connected to the study of loss landscapes and their geometry. Here, we introduce concepts from the loss landscape literature that inform contemporary approaches to model merging.

\subsection{Model Spaces}

Deep learning literature deals with several \enquote{spaces} over which we can analyze models. For clarity, we define these spaces here. \textit{Parameter space} or \textit{weight space} refers to the space of all model parameters. This is generally the domain of model merging techniques that seek to analyze models' weights and alter behavior via interventions on those weights. \textit{Activation space} refers to the space of all model activations. That is the set of possible activation values in each neuron in a particular deep learning model. This is generally the domain of interpretability techniques that attribute semantic behaviors in models to subnetworks of neurons using activation analyses---for example, measuring the correlation between activations in a subnetwork and the emergence of a behavior of interest.

\subsection{Loss Landscapes}
\label{sec:bg_loss_landscapes}

The \textit{loss landscape} (also referred to as the \textit{objective landscape}) of a deep learning model is the high-dimensional surface induced by the loss function when evaluated over the domain of all possible parameter values \cite{li2018visualizing}. \textit{Loss landscape geometry} refers to the geometric structure of the loss landscape, including the location of local minima and their relation to one another. A key concept is the idea of an \textit{objective mode}---a particular local minimum of the loss landscape. An objective mode lies in an \textit{objective basin} (or \textit{loss basin}), which is comprised of the objective mode and the surrounding locally convex region of solutions. Model merging methods must account for the structure of the loss landscape to synthesize performant solutions from existing models. Therefore, studying model training behavior and loss landscape geometry is crucial to developing effective model merging techniques. Below, we review fundamental loss landscape concepts.

\subsubsection{Mode Connectivity}

Consider a given path, $p(\theta_a, \theta_b, \lambda)$, between two models $\theta_a$ and $\theta_b$ parameterized by $\lambda \in [0,1]$. We will refer to $\theta_a$ and $\theta_b$ as the \textit{endpoints} of the path. The \textit{loss barrier}%
, defined below in \autoref{eq:loss_barrier}, on the path $p$ is defined as the maximum increase in loss along $p$ over the average loss of both endpoints. 

\begin{equation}
\label{eq:loss_barrier}
    \mathcal{L}_{p} = \max_{\lambda \in [0,1]} \mathcal{L}(p(\theta_a, \theta_b, \lambda)) - \frac{1}{2}(\mathcal{L}(\theta_a) + \mathcal{L}( \theta_b))
\end{equation}

Two models are said to exhibit \textit{Mode Connectivity} if there exists a path $p$ between them where the loss barrier is approximately zero \cite{draxler2018essentially, frankle2020linear, ainsworth2022git}. \textit{Linear Mode Connectivity}~(LMC) \cite{nagarajan2019uniform, frankle2020linear} is a stronger condition asserting that two modes can be connected by a straight line. Linear connectivity is of special interest as linearly connected solutions may lie in the same convex objective basin. Mode-connecting paths reveal the geometric structure of the loss landscape and allow us to find diverse but equally performant solutions between two known models \cite{juneja2022linear}.


In practice, we can discover paths of low loss between distinct neural networks via optimization (e.g., by parameterizing a Bézier curve) \cite{garipov2018loss, tatro2020optimizing, draxler2018essentially, tan2023geodesic, qin2022exploring}. Surprisingly, solutions are often connected by simple curves, and even linear mode connectivity is commonplace \cite{garipov2018loss, frankle2020linear}. Recent work has exploited this fact to ensemble networks that are close in the objective landscape, indicating that despite lying close together, sufficient diversity exists between these solutions to provide an advantage from ensembling~\cite{garipov2018loss}. Some empirical evidence suggests that geodesics between neural networks in the distribution space induced by the Fisher-Rao metric may correspond to paths between their parameters in objective space~\cite{tan2023geodesic}. This finding 
links the notions of mode connectivity and interpolating between distributions. Additionally, it has been shown that linear interpolation between the parameters of two linearly connected models corresponds to linear interpolation between their activation maps \cite{zhou2023going}. This property holds even when the endpoints of interpolation are two networks fine-tuned on distinct tasks, provided that they stem from the same pre-trained checkpoint (see \autoref{sec:training_trajectories}) \cite{zhou2024emergence}. 

\input{graphics/tex/fig_permutation_symmetry}

\subsubsection{Permutation Symmetry}
\label{bg_permutation_symmetry}

Neural networks inherently have components that are invariant to permutations. For example, in a single (dense) linear layer of neurons, any permutation in the order of neurons would still result in the same activations for that layer (see \autoref{fig:permutation_symmetry}). Therefore, we can consider any neural network to be a member of an equivalence class of other neural networks whose units are permuted but whose activations are the same. 
This permutation-invariance property has led some to conjecture that all solutions to SGD can be linearly connected if the natural permutation invariance of neural networks is accounted for \cite{entezari2021role}. This conjecture has spurred further investigation into the permutation symmetries inherent in the layers of neural networks and attempts to resolve them to achieve LMC between different solutions. 

Several theoretical results demonstrate that the objective landscapes of over-parameterized models are dominated by permutation-equivalent objective modes and that valleys in the loss landscape can be connected through permutation points \cite{simsek2021geometry, brea2019weight}. Additionally, many merging works provide substantial empirical evidence that aligning neurons by learning appropriate permutations in their order reveals linearly connected paths between models (see \autoref{permutation_alignment})
\cite{ainsworth2022git,singh2020model, li2024training, tatro2020optimizing, wang2020federated}. This property holds even when models are trained on slightly differing data distributions \cite{ainsworth2022git, singh2020model}.  Models trained on similar underlying distributions have been observed to learn weight matrices with similar distributions of singular values but differing singular directions. This observation has led some to propose that the success of neuron alignment methods can be explained by the resulting alignment of dominant singular directions in the weight matrices by learned permutations of the order of neurons (see \autoref{sec:neuron_alignment_limitations}) \cite{ito2024analysis}. This observation suggests we can meaningfully associate task-specific performance with just a few singular directions in parameter space \cite{ito2024analysis, vergara2024eigenpruning}.

\subsubsection{Training Trajectories}
\label{sec:training_trajectories}

The position of a neural network in the objective landscape is determined by the interaction between loss landscape geometry and gradient descent. Therefore, understanding \textit{training trajectories} is paramount in model merging. A training trajectory is a sequence of points, each representing the parameters of a neural network, that evolves under some optimization algorithm, such as SGD, during the training process. Numerous efforts have been made to characterize how training proceeds from distinct initialization points under SGD and how this relates to the underlying objective landscape. Empirically, researchers have observed that SGD is initially unstable but soon reaches a \enquote{stable point}. After this point, subsequent fine-tuning causes networks to converge to the same loss basin even under different forms of SGD noise (e.g., batch order) \cite{nakhodnov2022loss, frankle2020linear}. Some work has also demonstrated the converse, that deep neural networks with different random initializations learn functions with distinct properties and lie in distinct regions~\cite{fort2020deep}. 
These works all support the conclusion that neural networks that share a significant portion of their training trajectory tend to lie in the same linearly connected region of the objective landscape (see \autoref{sec:mode_determinism}). Conversely, those that are initialized separately tend to lie in distinct regions of the objective landscape. A corollary of this finding is that models in the same objective basin generally do \textit{not} require any form of neuron alignment for merging \cite{jin2022dataless}. 

%% file: graphics/tex/fig_permutation_symmetry.tex
\begin{figure}[ht!]
    \centering
    \includegraphics[width=1.0\linewidth]{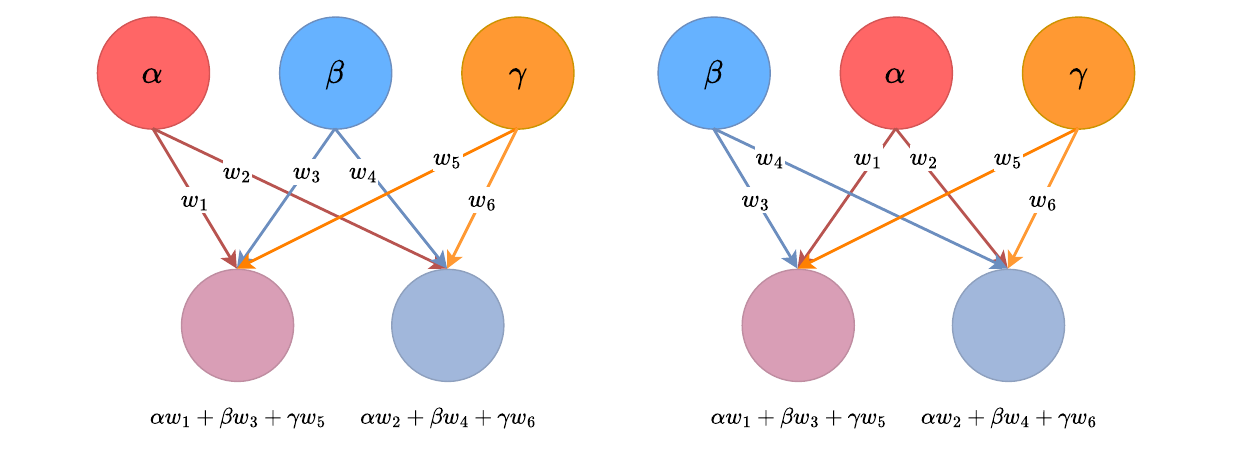}
    \caption{\textbf{Permutation Symmetry}: An illustration of symmetry in dense linear layers. Changing the order of two neurons does not change the resulting activations.}
    \label{fig:permutation_symmetry}
\end{figure}

%% file: sections/3_survey.tex
\section{Ensembling}
\label{sec:ensembling}

Ensembling is the most widely applied and reliable method to combine knowledge between models \cite{mienye2022survey}. Ensemble techniques combine the output logits of several models or compose their layers rather than combining their parameters (using, say, an average) to produce a single merged model. This means that rather than producing one model from several, these techniques apply several models to a single inference task; this makes their deployment in the contemporary billion or trillion parameter regime difficult. Despite the fact that ensemble techniques are not true model merging techniques in the way that we define them (in the sense that they do not actually combine model parameters), ensemble techniques are precursors to much of the model merging literature, therefore, we review several prominent ensembling techniques below to frame our subsequent discussion of weight aggregation and neuron alignment.


\subsection{Overview}

Ensembling combines the predictions (as opposed to parameters) from several trained models to enhance accuracy at inference-time. Ensembles compute a weighted combination of output predictions from each model in the ensemble to form a prediction. The success of ensemble methods is often attributed to a reduction in model bias and variance that occurs when predictions are averaged from several statistical models~\cite{fort2020deep, mienye2022survey}. Ensembling is closely related to Bayesian Model Averaging techniques~\cite{izmailov2018averaging}. In classic Bayesian Model Averaging, the final prediction is a weighted average of the predictions of each of our models, weighted by the posterior probability of each model~\cite{hoeting1999bayesian}.

For neural networks, the posterior probabilities are calculated over different weight settings:

\begin{equation}
    p\left(y \mid x, D\right) 
    = \int 
        p\left(y \mid x, \theta\right)\; 
        p\left(\theta \mid D\right)\, d \theta
\end{equation}

\noindent 
where $y$ is the data label, $x$ the input data, $D$ the training data, and $\theta$ the neural network parameters. Integrating over all weight settings for a neural network is intractable; however, many works observe that neural networks consistently converge to solutions lying in large, flat basins in the loss landscape~\cite{izmailov2018averaging, huang2020understanding, chaudhari2019entropy, hochreiter1997flat, keskar2016large}. One explanation for this behavior is that these flat areas occupy large volumes in parameter space, and thus, stochastic gradient descent is exponentially more likely to land in these regions \cite{huang2020understanding}. From a Bayesian perspective, these large, flat areas in parameter space have high probability density, $p\left(\theta \mid D\right )$, in the integral above. Therefore, an ensemble where we take the average predictions of several trained neural networks \cite{lakshminarayanan2017simple} can be thought of as an approximation of a Bayesian calculation of the posterior probability of $y$, where one estimates the integral above by summation over a few parameter settings with high probability under the posterior.

\subsection{Ensemble Techniques}

Despite the reliability of ensemble methods, they can be impractical because they require training and deploying several deep neural networks simultaneously. Some methods attempt to reduce the cost of obtaining several models to be used as ensemble members. \textit{Snapshot Ensembles} 
cycle through learning rates during training to produce several diverse models at various minima within a single training run \cite{huang2017snapshot}. Another approach is to exploit the local loss geometry around an existing model to discover useful ensemble members; for example, \textit{Fast Geometric Ensembling}~(FGE) \cite{garipov2018loss} samples new models from within the same objective basin as an existing model. This idea exploits the linear connectivity within an objective basin but has since been extended to exploit more complex forms of Mode Connectivity in the loss landscape. \citet{benton2021loss} develop a method that finds whole simplices of low loss in the objective landscape that connect several potential solutions. This method closely approximates a classic Bayesian model average where the posterior is uniformly distributed over a simplex. 

In addition to algorithmic advancements, recent increases in the number of freely available models have also reduced the cost of constructing ensembles. More pressing is the issue of runtime and memory efficiency, each of which scales linearly with the number of models in the ensemble. \textit{Mixture-of-Experts}~(MoE) routines reduce the inference cost associated with ensembling to that of a single model \cite{shazeer2017outrageously, masoudnia2014mixture}. Rather than averaging the predictions of each ensemble member, these methods learn to compose components---at the granularity of whole networks, layers, or individual neurons---from each model to compute the final output prediction. This amounts to learning how to construct a new neural network for each input presented. Traditional MoE learns a routing network which then selects $k$ of the ensemble members (\enquote{experts}) and computes a linear combination of their outputs to form the final prediction logits. This routing procedure was simplified by \citet{fedus2022switch} to produce \textit{Switch Transformers} which demonstrate that even using just a single expert per input token is sufficient for strong performance.

\subsection{Limitations of Ensembling}

Ensembles of several models have become much more straightforward to obtain thanks to advancements in our understanding of the loss landscape~\cite{garipov2018loss, benton2021loss, huang2017snapshot}. Still, ensemble techniques suffer from poor scaling in terms of runtime and memory costs, as we must use multiple models for inference. MoE methods are a boon for use cases where inference cost is a limiting factor; however, they also suffer from several drawbacks. While the effective parameter count for a given input in an MoE model is lower, the total number of parameters in the mixture may be the same or even greater \cite{pan2024dense}. As individual \enquote{experts} in the mixture specialize in just a few prediction domains, we must train more parameters in total to achieve strong performance across all domains. Consequently, training and serving MoE models requires constantly exchanging large numbers of parameters from memory. This requires a significant investment in complex infrastructure to serve models of billions or trillions of parameters. MoE models also exhibit decreased sample efficiency during training: each parameter is not trained during every batch, resulting in increased training costs. Though these methods offer exciting avenues for scaling and deploying ensemble models more efficiently, model merging offers the potential for equivalent performance at a significant cost reduction. 

\input{graphics/tex/fig_mode_convexity}

\section{Weight Aggregation}
\label{sec:weight_averaging}

\input{sections/weight_agg_table}

The fundamental technique underlying most model merging methods is weight aggregation. Given two or more models of the same architecture, one can compute an average or convex combination of the values for each corresponding parameter to obtain a set of new parameter values that comprise the merged model. These techniques are predicated on the principles of Bayesian Model Averaging, seeking to reduce estimation error from individual model bias and variance by combining several models for inference \cite{hinne2020conceptual}. Though weight aggregation methods show promise compared to ensemble methods, they also introduce several challenges. Weight aggregation methods usually introduce additional hyperparameters that must be set via trial-and-error or by some heuristic. Additionally, their limitations are not fully understood: while some characterizations of their failure cases can be extrapolated from the literature (see \autoref{permutation_alignment} below), more context on loss landscape geometry is necessary to fully understand their modes of failure. Below, we discuss several modes of weight aggregation, including simple averaging, weight steering via task arithmetic, data-aware weightings, and information-theoretic weightings. \autoref{tab:weight_agg_overview} lists the weight aggregation techniques we cover in this section and enumerates some of their distinguishing features.

\subsection{Simple Averaging}
\label{sec:simple_averaging}

The most straightforward method of merging model parameters is to take a simple average over each corresponding parameter between two or more models, see \autoref{eq:simple_avg}. 
This technique is also employed in federated settings as \textit{FedSGD} \cite{mcmahan2017communication}.
However, simple averaging does not consider each model's relevance to the task at hand. 
This method also consistently fails if the training distributions or the training trajectories between the two models have diverged \cite{singh2020model, jin2022dataless, ainsworth2022git}. 
Despite this limitation, simple averaging is particularly useful in the \enquote{pre-train and fine-tune} model deployment paradigm, where models naturally share an architecture and are predisposed towards similar representations through pre-training~\cite{zhou2023going, zhou2024emergence, erhan2010does}. 
LLMs are a good example of this kind of deployment. 

\begin{equation}
    \theta^\star = \frac{1}{N} \sum_{i} \theta_i
\label{eq:simple_avg}
\end{equation}

A host of methods attempt to improve performance by using variations of simple averaging. \textit{Stochastic Weight Averaging}~(SWA)~\cite{izmailov2018averaging} averages checkpoints from a single model at different points in its training trajectory, demonstrating comparable performance to FGE \cite{garipov2018loss}. 
\textit{Model Soups}~\cite{wortsman2022model} are simple averages of fine-tuned models of the same architecture trained with different hyperparameters. Model soups often enhance \textit{Out-Of-Distribution}~(OOD) and zero-shot performance on downstream tasks. 
\textit{Sparse soups}~\cite{zimmer2023sparse} reduce the inference costs associated with model soup methods by enforcing a sparsity constraint via an iterative prune-retrain fine-tuning technique inspired by \textit{Iterative Magnitude Pruning}~(IMP)~\cite{tanaka2020pruning}. Similarly, the
\textit{SWAMP}~\cite{choi2023swamp} method iteratively performs IMP and simple weight averaging between multiple fine-tuned models to produce a final sparse model. These works support contemporary observations that models closer to the center of a group of fine-tuned models stemming from the same pre-trained model are more generalizable and robust \cite{gueta2023knowledge, jang2024model}. \citet{wortsman2021learning} demonstrate that several networks can be trained in tandem such that they form a linearly connected subspace of solutions. 
The midpoint of this subspace approaches the accuracy of an ensemble of independently trained models and demonstrates robustness to label noise. \textit{Model Stocks}~\cite{jang2024model} take advantage of this observation by estimating a \enquote{central} network using only two fine-tuned models, significantly reducing the training effort required to construct model soups.

\subsection{Weight Steering} 
\label{sec:param_nudging}

\textit{Task Arithmetic}~\cite{ilharco2022editing} is a technique for manipulating model weights to enhance task-specific and OOD performance. A task vector---defined as the difference between the pre-trained and fine-tuned model weights for a task, see \autoref{eq:task_vector}---can be thought of as the direction in weight space that the pre-trained model must travel to achieve better accuracy on that particular task. 

\begin{equation}
    \Delta_\text{task} = \theta_\text{task} - \theta_\text{pre}
\label{eq:task_vector}
\end{equation}

Interestingly, combining task vectors via simple arithmetic operations (addition and subtraction) provides predictable control over model behavior. For example, subtracting a task vector from the pre-trained weights reduces performance, while adding the same task vector improves performance on the corresponding task; adding two task vectors for distinct tasks to the pre-trained model weights enhances the resulting model's performance on both tasks.
This property is exploited by
\textit{Model Breadcrumbs}~\cite{davari2023model}, a method that takes task vectors from several fine-tuned models, masks outlier directions, and combines them via addition into a single parameter update for the pre-trained checkpoint. Steering weights using task vectors can enable good performance in multiple tasks without degrading performance in any individual task significantly, particularly in larger models \cite{ilharco2022patching}.
\textit{TIES-merging}~\cite{yadav2024ties} and \textit{DARE}~\cite{yu2024language} improve upon Task Arithmetic by eliminating redundancies between task vectors, resulting in improved performance---particularly in the multi-task regime. It is worth noting the connection between these methods and parameter-efficient fine-tuning schemes such as LoRA~\cite{hu2021lora, hayou2024lora+, liu2024dora} and IA3~\cite{liu2022few}. LoRA trains low-rank decompositions of weight matrices during fine-tuning that are added to the model weights for inference, while IA3 trains per-task scaling vectors in each layer that rescale the activations in that layer---equivalent to rescaling the weight matrix along corresponding directions in weight space. It is common to \enquote{swap} fine-tuning weights for different tasks by adding and subtracting out LoRA weights---a procedure analogous to how we use task vectors. \textit{LoReFT} \cite{wu2024reft}, inspired by LoRA, uses Distributed Alignment Search (DAS) \cite{geiger2024finding, wu2024interpretability} to find task-specific directions in which to update low-rank fine-tuning weights. 

The success of weight steering methods implies that, within a single objective basin, we can meaningfully assign semantic properties to directions in weight space. Furthermore, we can leverage the linearly connected nature of local minima to smoothly interpolate between task-optimal parameter settings within a basin---even for multiple tasks \cite{tang2024parameterefficientmultitaskmodel}. This raises the question: can we improve the performance of model merging techniques by calibrating the contribution of task-specific parameters to the final merged model?

\input{graphics/tex/fig_regmean}

\subsection{Data-Aware Weightings}
\label{sec:heuristic_weightings}

\textit{AdaMerging}~\cite{yang2023adamerging, zhao2024adamergex} builds upon Task Arithmetic, learning merging coefficients for task vectors adaptively by minimizing prediction entropy over unlabeled test data. The idea of minimizing prediction entropy originates in test-time adaptation schemes \cite{wang2020tent, liang2024comprehensive} and has been related to finding large, flat basins associated with enhanced generalization capacity \cite{chaudhari2019entropy}. AdaMerging demonstrates better multi-task performance than any other merging method in this section, approaching that of an ensemble. 
However, it still falls short of the accuracy of a monolithic multi-task model.

Other approaches analyze model activations to determine the importance of particular network components to the final merged model. \textit{RegMean}~\cite{jin2022dataless} merges models trained on separate datasets by computing weight matrices that are the least squares fit to their activation patterns. That is, it attempts to find a weight matrix in each layer that, when presented with an input from the dataset $X_i$ corresponding to the source model $\theta_i$, produces activations with minimal $L^2$ distance from those of $\theta_i$. Rather than attempting to match activation maps, \textit{ZipIt!}~\cite{stoica2023zipit} seeks to merge only redundant model components whose activations are highly correlated and preserves task-specific components. Intuitively, this prevents destructive interference between task-specific parameters. 
ZipIt!\ demonstrates strong performance in multi-task settings---even when models are trained independently. \textit{Evolutionary Model Merging} is a similar technique that uses an evolutionary algorithm to learn a policy for whether to compose layers from each source model in sequence or to learn merging weights for each parameter in a given layer~\cite{akiba2024evolutionary}.

Some applications require the training of neural networks on data held by edge computing devices without moving data to other locations~\cite{xu2021federated, hudson2022smart, hudson2021framework,devaraj2024ruralai}.
A common paradigm for such contexts is \textit{Federated Learning}~(FL) where models are trained separately on edge devices and then merged at a central server using FedSGD \cite{mcmahan2017communication}, a simple average over model parameters. 
Federated Averaging (FedAvg)~\cite{mcmahan2017communication, sun2022decentralized} generalizes this technique to deal with non-IID data distributions, which degrade the efficacy of simple averaging.
FedAvg weights the model parameters returned by each device by the number of training data samples used to update the model parameters, see \autoref{eq:fedavg}. Empirically, this heuristic handles non-IID training well~\cite{mcmahan2017communication}. Under IID distributions, FedAvg is equivalent to FedSGD.
FedAvg is defined formally as:

\begin{equation}
\theta_{t+1}^{\star} = \sum_{k=1}^{K} \frac{n_{k}}{n} \theta_{t}^{k}
    \label{eq:fedavg}
\end{equation}

\noindent
where $K$ is the number of devices in the federated setting, $n_{k}$ is the number of data samples at device~$k$, $n=\sum_{k} n_k$, $\theta_{t+1}^{k}$ is the parameters of device $k$ in training round $t$, and $\theta_{t+1}^{\star}$ is the updated global model which merges all the models locally-updated on the individual devices.


\subsection{Information Theoretic Weightings}

\textit{Fisher Weighted Averaging}~\cite{matena2022merging} uses the Fisher information of model parameters as weights for a weighted average of parameters. Due to the massive size of the Fisher matrix for neural networks, they use a diagonal approximation to make this method tractable. 
\textit{MaTS} \cite{tam2023merging} improves this method by producing a block-diagonal approximation of the Fisher matrix via Kronecker Factorization \cite{martens2015optimizing, grosse2016kronecker, ba2017distributed} and solving the resulting linear system of equations for the optimal merged parameter values using the Conjugate Gradient method \cite{hestenes1952methods}.

\subsection{Limitations of Weight Aggregation}

Weight aggregation methods 
have similar advantages to ensembles, such as enhanced generalization, but with a significantly reduced cost. However, they are only applicable and well-defined for merging models of the same architecture. This property precludes their use for more general forms of knowledge transfer than those studied in this section. Another drawback is that their performance degrades as models diverge in their training trajectory. The studies surveyed generally attempt to merge fine-tuned models from the same pre-trained checkpoint; attempts to perform weight aggregation naively on independently trained models are usually unsuccessful~\cite{yamada2023revisiting}. Similarly, weight aggregation between models trained on tasks with significantly different data distributions tends to be unsuccessful. An exception is ZipIt!~\cite{stoica2023zipit}, which gracefully handles divergence in training trajectory and merging between models with different layer sizes---this is because it only merges redundant neurons in each layer. ZipIt! also circumvents issues regarding neuron alignment between layers, as they do not naively merge neurons.

The tendency for weight aggregation methods to fail when training conditions differ may be explained by underlying permutation misalignment between neural networks \cite{frankle2020linear, simsek2021geometry} or by significant differences in the underlying learned features \cite{ito2024analysis, yamada2023revisiting}, which make it infeasible to directly combine corresponding parameters. Still, the constraints necessary for weight aggregation to succeed illustrate interesting properties of model training behavior. The ability to interpolate smoothly between fine-tuned models originating from the same checkpoint without significant increases in loss along this path suggests that those models lie in the same linearly connected region, this notion of \textit{Mode Convexity} is illustrated in \autoref{fig:mode_convexity}. The predictable success of interpolating between fine-tuned models aligns with empirical data indicating that at a certain point in a model's training trajectory, the final linearly connected objective basin to which they will converge is pre-determined---even if the training distribution is changed after this point in the trajectory \cite{frankle2020linear}. 

A linear path between weights has also been shown to reflect a linear path between learned features in some scenarios \cite{zhou2023going}, particularly under the pre-training and fine-tuning paradigm \cite{zhou2024cross}. This result may explain why some aggregation procedures might succeed without accounting for permutation invariance. Moreover, the success of weight steering methods indicates that---at least within an objective basin---we can accurately assign directions in parameter space to task-specific behavior. This property has also been noted by \citet{gueta2023knowledge}, who observe a clustering in parameter space for LLMs fine-tuned from the same pre-trained checkpoint on semantically similar tasks.

\input{graphics/tex/fig_mode_connectivity}

\section{Neuron Alignment} 
\label{permutation_alignment}

\input{sections/neuron_alignment_table}

As opposed to naive weight aggregation schemes, neuron alignment techniques reduce potential degradation in the resulting merged model by first aligning the neurons of each source model and then performing weight aggregation. This prevents weight aggregation from destroying important features from source models, which may occur if the source models are permuted relative to one another as described in \autoref{bg_permutation_symmetry}. Neuron alignment schemes outperform naive weight aggregation when models are trained independently or otherwise lie in different regions of the objective landscape, often revealing linear paths between distinct modes. \autoref{tab:neuron_alignment_overview} lists the works we review in this section along with some of their high level features.

\subsection{Learning Permutation Matrices}
\label{learning_perm_matrices}

\citet{entezari2021role} conjecture that all objective modes are linearly connected if one accounts for permutation symmetries between modes, this notion of \textit{Mode Connectivity} is illustrated in \autoref{fig:mode_connectivity}. To this end, various neuron alignment techniques attempt to explicitly determine a permutation matrix that can align the neurons of one model to those of another. 
\citet{ainsworth2022git} present two methods for aligning neurons via permutation matrices: activation matching and weight matching. 
Activation matching aligns each corresponding layer between two source models by finding a permutation matrix that minimizes the $L^2$ distance of their activation patterns. 
Weight matching finds a permutation matrix that attempts to maximize the Frobenius product between weight matrices. 
Intuitively, weight matching aligns weights that are similar to one another and produce similar activations as a result. 
These methods are set up as a sum-of-bilinear-assignments problem and solved using an iterative approximation algorithm. 
Both methods are competitive with their third proposed method, which is to learn a permutation matrix directly via a straight-through-estimator (STE). This method updates the weights of the second source model, $\theta_b$, to directly minimize the loss barrier (\autoref{eq:loss_barrier}), projecting the result to the nearest realizable permutation of $\theta_b$'s parameters. 
Notably, the performance of all three methods degrades as model width decreases and similarly worsens for models in the initial stages of training. 

\citet{tatro2020optimizing} relate the neuron alignment problem to the assignment problem \cite{burkard1999linear} and optimal transport. They align neurons to maximize the correlation between their pre-activations and solve the resulting alignment problem using the Hungarian Algorithm \cite{kuhn1955hungarian}. They also provide theoretical results indicating that such alignment tightens the upper bound on the loss of interpolated models. Interestingly, this method also performs better on wider model architectures. \textit{Probabilistic Federated Neural Matching} (PFNM) \cite{yurochkin2019bayesian} seeks to align the parameters of edge models to a global model in the federated setting using a Bayesian nonparametric model. \textit{Federated Learning with Matched Averaging} \cite{wang2020federated} takes inspiration from PFNM and FedAvg by using the Hungarian Algorithm to align corresponding layers, reducing the communication burden associated with merging in the federated setting. 

\subsection{Learning Alignment Maps}

Rather than constraining themselves to learning permutation matrices, some methods instead seek to learn unconstrained \enquote{soft} maps that enable neuron alignment for merging. \textit{Optimal Transport Fusion} (OTFusion) \cite{singh2020model} poses the model merging problem as an optimal transport problem and aligns neurons based on activation and weight similarities using an exact optimal transport solver, requiring only a single pass over all layers as opposed to the other iterative neuron alignment methods presented in \autoref{learning_perm_matrices}. The optimal transport framework allows for merging between models with differing layer sizes. While previous methods explicitly constrain optimization to return permutation matrices, OTFusion imposes no such constraint. Instead, it finds a Wasserstein barycenter between the distributions encoded by each source model. This amounts to optimization over the Birkhoff polytope \cite{ziegler2012lectures} in $n$ dimensions whose vertices are the traditional permutation matrices in $S_n$ (if the model layer sizes are equivalent). In this regard, OTFusion is a generalization of prior methods that relate neuron alignment to the discrete assignment problem. Once again, we note that OTFusion's performance improves with increased model width. 

OTFusion has since been extended to handle transformer architectures~\cite{imfeld2023transformer}---specifically dealing with residual connections, multi-head attention, and layer normalization---as well as to align neurons across layers in heterogenous architectures (as opposed to only within corresponding layers)~\cite{nguyen2023cross}. 
Despite their good performance, even aligned neural networks sometimes demonstrate high loss barriers on the path between them. \citet{jordan2022repair} attribute this problem to a reduction in the variance of hidden layer activations during interpolation and propose an intermediate scaling layer that improves the performance of models merged using neuron alignment techniques, reducing the loss barrier between models anywhere from 60-100\%.

\subsection{Limitations of Neuron Alignment}
\label{sec:neuron_alignment_limitations}

Merging procedures combined with neuron alignment seem to resolve issues raised by prior work associated with merging models from independent training trajectories. Permutation-based methods also empirically reinforce Entezari et al.'s conjecture~\cite{entezari2021role} that many, if not all, objective modes are linearly connected if one accounts for permutation symmetries. One drawback of the current state-of-the-art for neuron alignment is that these methods are often forced to perform greedy layer-wise alignment for tractability when, in principle, permutation symmetries may exist across layers as well. Therefore, there may be a set of permutation invariances that these methods do not account for---as well as other yet undiscovered forms of misalignment between neurons. 

Additionally, the issue of merging models trained on different tasks remains. STE \cite{ainsworth2022git} and AdaMerge \cite{yang2023adamerging} both show promise in this regard but require data samples and backpropagation---purely manipulating weight matrices seems insufficient. \citet{yamada2023revisiting} explore this problem, 
empirically demonstrating that merging performance degrades as the training distributions of source models diverge. They conjecture that the growing differences in loss geometry that arise as datasets diverge require optimization over samples from a mixed-task dataset. They use Dataset Condensation \cite{zhao2020dataset} to cheaply construct mixed datasets and optimize merged parameters by using the STE method, with promising results (albeit only on MNIST and Fashion-MNIST, both known to be relatively \enquote{easy} datasets \cite{frankle2020linear}). 

\citet{ito2024analysis} investigate the success of neuron alignment methods further by analyzing the net effect of neuron alignment methods 
on the properties of weight matrices. Proponents of permutation-based methods often paint an intuitive picture of models being transported into the same convex objective basin, evoking an image of reduced $L^2$ distance between weights. However, their investigation shows a poor correlation between neuron alignment and $L^2$ distance. Instead, the benefit of neuron alignment seems to be predicated on the alignment of dominant singular directions in the weight matrices of each layer. They also show that models independently trained on similar tasks tend to show a similar distribution of singular values. This suggests that difficulties in merging models trained independently on the same task may be explained by differences in their singular directions rather than by fundamental differences in learned features. \citet{jin2022dataless} indicate the converse: that models stemming from the same pre-trained model (i.e., that lie in the same objective basin) show no evidence of permutation between weights. This implies that permutation-based procedures are only necessary when source models are trained independently. \citet{ito2024analysis} also explain the correlation between greater merging success and model width by the fact that as model width increases, the distribution of dominant singular values becomes more concentrated. The existence of fewer dominant feature directions to align makes the alignment process more manageable (see \autoref{sec:model_size}).

These studies further indicate that differences in learned features between models trained on different tasks may render procedures based purely on weight manipulation ineffective. Merging across tasks may require optimization via backpropagation or avoiding merging task-specific components altogether, as in ZipIt!~\cite{stoica2023zipit}.

%% file: graphics/tex/fig_mode_convexity.tex
\begin{figure}[t]
    \includegraphics[width=0.5\linewidth]{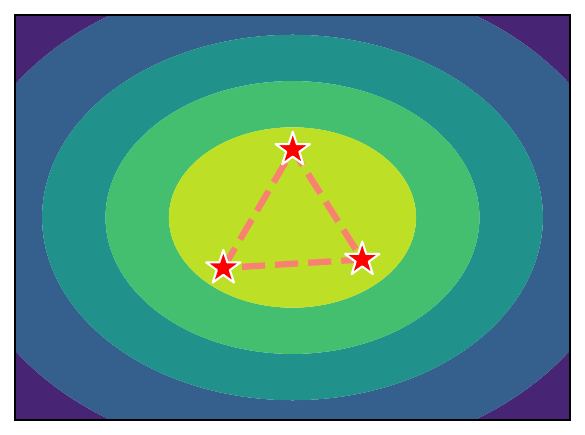}
    \caption{\textbf{Mode Convexity:} Within an objective basin, one can linearly interpolate between known solutions to discover equally performant models with diverse behaviors.}
    \label{fig:mode_convexity}
\end{figure}

%% file: sections/weight_agg_table.tex
\begin{table}[H]
\centering
\caption{Overview of Weight Aggregation techniques}
\label{tab:weight_agg_overview}
\begin{tabular}{p{0.3\textwidth}cccc} 
\toprule
Technique & Simple Average & Weighted Average & Requires Data & Requires Labels \\
\midrule
FedSGD~\cite{mcmahan2017communication} & \yes &  \no & \no & \no \\
Stochastic Weighted Averaging~\cite{izmailov2018averaging} & \yes & \no & \yes & \yes        \\
Model Soup~\cite{wortsman2022model} & \yes & \no & \yes & \yes \\
Sparse Soup~\cite{zimmer2023sparse} & \yes & \no & \yes & \yes \\
SWAMP~\cite{choi2023swamp} & \yes & \no & \yes & \yes \\
Task Arithmetic~\cite{ilharco2022editing} & \yes & \no & \no & \no \\
Model Breadcrumbs~\cite{davari2023model} & \yes & \no & \no & \no \\
TIES-merging~\cite{yadav2024ties} & \yes & \no & \no & \no \\
DARE~\cite{yu2024language} & \yes & \no & \no & \no \\
FedAvg~\cite{sun2022decentralized} & \no & \yes & \no & \no \\
AdaMerging~\cite{yang2023adamerging} & \no & \yes & \yes & \no \\
RegMean~\cite{jin2022dataless} & \no & \yes & \yes & \no \\
ZipIt!~\cite{stoica2023zipit} & \yes & \no & \yes & \no \\
Evolutionary Model Merging~\cite{akiba2024evolutionary} & \no & \yes & \yes & \yes \\
Fisher Weighted Average~\cite{matena2022merging} & \no & \yes & \yes & \no \\
MaTS~\cite{tam2023merging} & \no & \yes & \yes & \no \\
\bottomrule
\end{tabular}
\end{table}

%% file: graphics/tex/fig_regmean.tex
\begin{figure}[t]
    \includegraphics[width=0.5\linewidth]{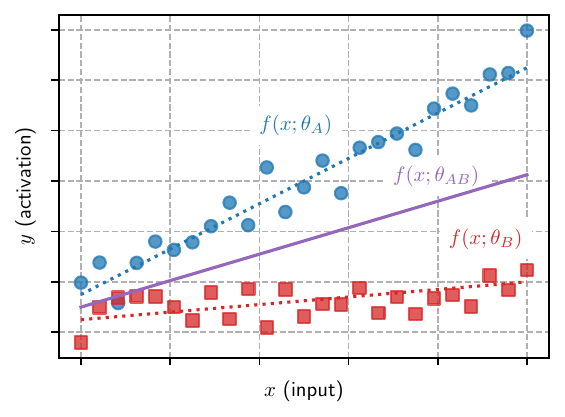}
    \caption{\textbf{RegMean} finds a new set of merged parameters that closely approximates the activation maps of the given source models in a manner analogous to least-squares regression.}
    \label{fig:regmean}
\end{figure}

%% file: graphics/tex/fig_mode_connectivity.tex
\begin{figure}[t]
    \includegraphics[width=0.5\linewidth]{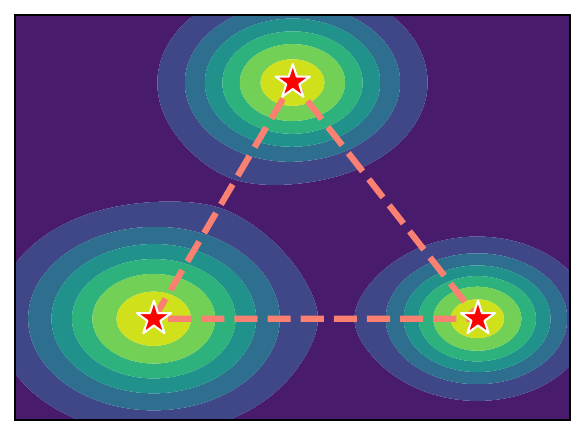}
    \caption{\textbf{Mode Connectivity:} Many objective basins are equivalent up to a permutation in neurons. Models in distinct objective basins can be \enquote{transported} close to one another by applying appropriate permutations to their units.}
    \label{fig:mode_connectivity}
\end{figure}

%% file: sections/neuron_alignment_table.tex
\begin{table}[H]
\caption{Overview of Neuron Alignment techniques}
\label{tab:neuron_alignment_overview}
\centering
\small
\begin{tabular}{
    p{0.3\textwidth}
    c
    c
    c
    c
} 
\toprule
    Technique & Finds Permutation Matrix & Finds \enquote{Soft} Mapping & Requires Data & Requires Labels \\
    \midrule
    Weight Matching~\cite{ainsworth2022git}
        & \yes{} & \no & \no & \no \\
    Heuristic Neuron Alignment~\cite{tatro2020optimizing} 
        & \yes & \no & \yes & \no \\
    Probabilistic Federated Neural Matching~\cite{yurochkin2019bayesian} 
        & \no & \no & \yes & \no \\
    FedMA~\cite{wang2020federated} 
        & \yes & \no & \yes & \no \\
    OTFusion~\cite{singh2020model} 
        & \no & \yes & \yes & \no \\
    Transformer OTFusion~\cite{imfeld2023transformer} 
        & \no & \yes & \yes & \no \\
\bottomrule
\end{tabular}
\end{table}

%% file: sections/4_insights.tex
\section{Insights into Training}
\label{sec:insights}

The overlap between the literature on model merging and loss landscape geometry 
yields several insights into the nature of loss landscapes and the evolution of model representations during training. Here, we present insights from the model merging literature into the underlying geometric structure of the loss landscape and the effects of model architecture on learned representations.

\subsection{Loss Landscape Macrophenomena}
\label{training_phenomena}

We outline macrophenomena consistently observed when model merging is performed. We study characterizations of these phenomena from investigations on loss landscapes and relate them to empirical data gathered through merging studies.

\subsubsection{Mode Convexity:}
Objective basins are locally convex, permitting one to travel a short distance and still encounter a meaningful neural network solution with similar behavior. More importantly, models with roughly equivalent or lower loss but meaningfully different prediction behavior exist in proximity to any given solution. 
\textit{Fast Geometric Ensembling}~(FGE)~\cite{garipov2018loss} implicitly takes advantage of this phenomenon to rapidly discover quality ensemble members after training only a single model and
\textit{Stochastic Weight Averaging}~(SWA)~\cite{izmailov2018averaging} exploits this phenomenon by averaging rather than ensembling several models from the same training trajectory. These studies suggest that local minima in the objective landscape contain models with sufficiently diverse behavior to gain a performance advantage by ensembling or merging them---despite their proximity in parameter space.


\input{graphics/tex/fig_mode_determinism}

\subsubsection{Mode Determinism:}
\label{sec:mode_determinism}

The final linearly connected objective basin to which deep learning models converge is determined early in the training process. In particular, models that share a significant portion of their training trajectory (e.g., models fine-tuned from the same pre-trained checkpoint) will converge to solutions in the same objective basin (see \autoref{fig:mode_determinism}). \citet{frankle2020linear} illustrate the tendency of models to become stable to sources of SGD noise (e.g., batch order, data augmentation) early in training, even observing that at a certain point in the training trajectory, models are determined to lie in the same linearly connected region. 
This assertion is corroborated by \citet{gueta2023knowledge}, who observe that fine-tuned language model checkpoints cluster loosely around their pre-trained ancestors, even when trained on semantically dissimilar tasks. SGD has been observed to occur in two phases: an initial period of instability, followed by a more stable, linear trajectory~\cite{frankle2020linear, frankle2018lottery, damian2022self, fort2019deep, ma2022beyond, cohen2021gradient}.
\citet{fort2020deep} show that this initially chaotic, eventually stable optimization behavior corresponds to the expected dynamics of the Neural Tangent Kernel---observing that within just 2-3 epochs, the final linearly connected region that a solution will lie in is already determined. 
Weight aggregation techniques implicitly use mode determinism; otherwise, they could not merge models using linear combinations of their parameters. Instead, we would expect barriers of high loss between models that were independently fine-tuned. 
Linearly connected models also demonstrate linearity in their features---that is, linear interpolations between their parameters correspond to linear interpolations between their feature maps \cite{zhou2023going, zhou2024cross}. 
This result implies that the nature of the representations that models will learn is also determined early in the training process. \citet{damian2022self} explain the stability in the later part of the training trajectory for neural networks by the tendency of SGD to implicitly regularize the sharpness of the loss landscape. They demonstrate the explicit presence of such a regularization term in the Taylor expansion of the SGD update equation. This result suggests that the observed stability in the training trajectory is a property of SGD itself.

\subsubsection{Mode Directedness:} 
Within a given objective basin, we can meaningfully associate directions in parameter space with task-specific behavior, which is depicted in \autoref{fig:mode_directedness}. This \enquote{directedness} is best illustrated by the success of methods such as Task Arithmetic~\cite{ilharco2022editing}, which explicitly construct task vectors that associate task-specific behavior with particular directions in parameter space. Follow-up work, such as  
\textit{TIES-merging}~\cite{yadav2024ties} and \textit{DARE}~\cite{yu2024language},
demonstrate that removing redundancies and low-magnitude differences between task vectors improves merging performance. This implies that just a few principal directions in parameter space can effectively account for task-specific behavior. \citet{gueta2023knowledge} demonstrate the clustering behavior of language models trained on similar tasks. They observe that models trained on the same dataset are tightly clustered in parameter space, whereas models trained on the same underlying task with different datasets form looser clusters. This notion of hierarchically organized task-specific areas in parameter space offers an explanation for the success of several standard routines in the contemporary deep learning landscape, including the pre-train fine-tune training paradigm and meta-learning procedures such as Model-Agnostic Meta-Learning \cite{finn2017model}. These procedures may move parameters into regions of space where effective parameter configurations for more specialized sub-tasks are abundant. While the existence of meaningful task-specific regions in parameter space has been proposed previously to motivate meta-learning approaches, the model merging literature justifies these conjectures with empirical evidence showing that this kind of organization is present at multiple scales.



\subsubsection{Mode Connectivity:} 
Even solutions found in distinct objective modes can be connected by low-loss paths that are comprised of equally performant solutions. 
Moreover, this connectivity between solutions in separate modes is commonplace, as is LMC, i.e., the presence of linear paths of low loss between known solutions in objective space. 
There may even be entire manifolds of low loss connecting equally performant solutions \cite{benton2021loss}. 
An accumulation of evidence suggests that models trained on similar tasks, even with different initializations, can be connected by paths of low loss if one accounts for the inherent permutation invariance of neural networks by appropriately aligning their parameters after training \cite{entezari2021role, ainsworth2022git, singh2020model, frankle2020linear, tatro2020optimizing}. 
Additionally, \citet{li2024training} showcase how constraining training of a network to a subspace or pruning prior to training reduces the number of inherent permutation symmetries and leads to LMC between solutions without further post-training alignment. This result implies that constraining neuron misalignment from arising can enable the discovery of connected paths between modes more easily.

\citet{simsek2021geometry} present several theoretical results suggesting that subspaces of permutation-equivalent objective basins corresponding to global minimizers dominate the loss landscape of heavily overparameterized models. This result suggests that vast overparameterization can lead to a more convex objective landscape. Under a quasiconvex objective landscape consisting of many permutation equivalent minima, model representations may generally be quite similar---even under different forms of SGD noise or initializations, they tend to converge to similar solutions modulo a permutation symmetry. This similarity would be even more prevalent as models are further overparameterized, offering an explanation for why merging methods are often more successful as model size increases.

\subsection{Implications for Model Training}

The direct comparison of model representations in merging studies allows us to derive information about the effects of model architecture and pre-training objectives on the development of learned representations. 
Here, we outline some of these findings.



\subsubsection{Model Size:} 
\label{sec:model_size}
Neuron alignment procedures struggle as model width decreases. \citet{ito2024analysis} investigate this phenomenon by studying the distribution of singular values in the weight matrices of source models. 
They conjecture that the effectiveness of neuron alignment techniques is predicated on their ability to preferentially align the singular directions corresponding to dominant singular values. 
They find that the proportion of dominant singular values decreases substantially as model width increases, suggesting that wider models are easier to align. 
\citet{nguyen2023cross} show empirical results suggesting that over-parameterized models propagate dominant principal components of their hidden representations at each layer. 
These results support the idea that a few dominant feature directions arise in over-parameterized models. \citet{aghajanyan2020intrinsic} measure the intrinsic dimensionality of language models and similarly conclude that LLMs form more compressed representations as they scale in size. 
These works 
suggest that larger models, despite their increased expressiveness, have fewer dominant directions in feature space. 

Some work has also attempted to clarify why over-parameterized models may display more amicable optimization behavior than smaller models. 
\citet{huang2020understanding} and \citet{li2018measuring} visualize the loss landscapes of models at various sizes and find that overparameterized models often converge to large, flat basins correlated with enhanced generalization capability \cite{izmailov2018averaging, huang2020understanding, chaudhari2019entropy, hochreiter1997flat, keskar2016large, dziugaite2017computing, li2020hessian} as opposed to small sharp minima. 
They explain this tendency by observing that such basins occupy far more volume in objective space as the dimensionality of a model increases. 
The increased volume of flat basins, coupled with the results presented by \citet{simsek2021geometry} indicating that the number of permutation-equivalent objective basins corresponding to global minimizers increases rapidly as models grow in size, may explain the relative ease with which large models train: the volume of objective space corresponding to good solutions grows rapidly with model size.


\input{graphics/tex/fig_mode_directedness}

\subsubsection{Pre-training Objective:}
Models with different pre-training objectives display different sensitivities to various merging techniques \cite{jin2022dataless, tam2023merging}. 
For example, transformer models in the T5 family \cite{raffel2020exploringt5} perform well even with only simple averaging; meanwhile, DeBERTa models~\cite{he2020deberta} consistently degrade in performance when merged, even when using state-of-the-art techniques \cite{jin2022dataless}. 
This discrepancy points to an interplay between the pre-training objective and the success of model merging procedures. 
One possible explanation for this difference is the effect of pre-training objective on training dynamics. For example, contrastive pre-training has been observed to coerce models to lie in flatter objective basins associated with better generalization and robustness guarantees \cite{fradkin2022robustness}. Longer pre-training procedures seem to predispose models toward representations that display linear connectivity and linear relationships between activations \cite{jin2022dataless,zhou2024cross}. Understanding this interaction is important for interpretability work that seeks to isolate task-specific behavior~\cite{meng2022locating, meng2022mass, wortsman2019discovering} or to disentangle and identify key directions in latent space~\cite{bricken2023towards}. 
Some work has explored the overlap between pre-training objective, loss landscape geometry, and adversarial robustness, even demonstrating how to prevent fine-tuning pre-trained models on restricted task domains~\cite{fradkin2022robustness, deng2024sophon}. Further work might consider the causal effects of pre-training objective on loss landscape geometry and relate it to resulting model robustness and performance in downstream tasks.

%% file: graphics/tex/fig_mode_determinism.tex
\begin{figure}[t]
    \includegraphics[width=0.75\linewidth]{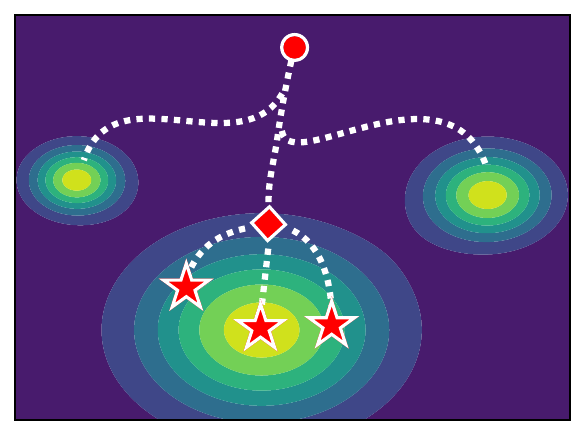}
    \caption{
    \textbf{Mode Determinism:} 
    As a training trajectory proceeds from the initialization point (pictured as the red circle here) its path may diverge to various minima in the loss landscape due to SGD noise (e.g., batch order) or changes in the underlying data distribution, but after a certain number of steps the trajectory of a model reaches a stable point (pictured as the red diamond). Any further fine-tuning from this point onwards, even on slightly different data distributions, will result in models lying in the same convex objective basin (red stars).}
    \label{fig:mode_determinism}
\end{figure}

%% file: graphics/tex/fig_mode_directedness.tex
\begin{figure}[t]
    \includegraphics[width=0.7\linewidth]{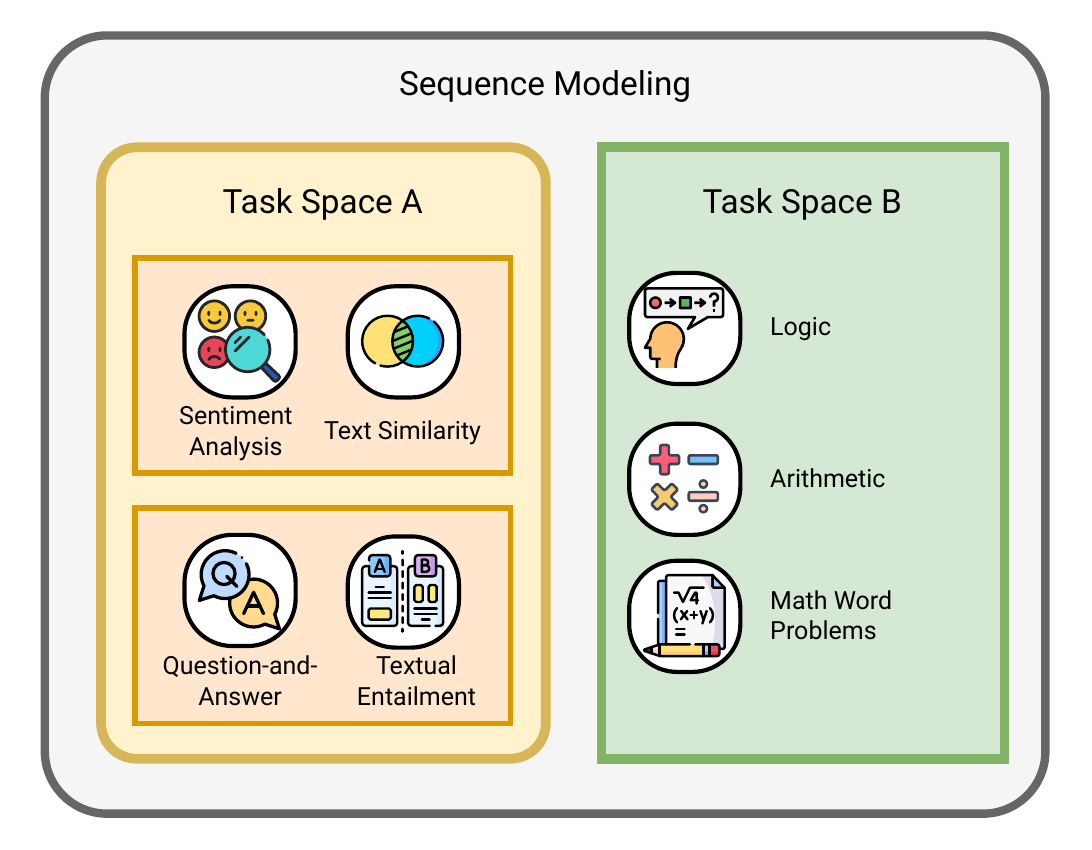}
    \caption{\textbf{Mode Directedness:} Optimal parameter settings for related tasks lie clustered together in parameter space; clusters become looser as tasks diverge in similarity. This suggests a hierarchical organization of the objective landscape with respect to semantically similar tasks. We use a handful of example natural language tasks here to illustrate this concept.}
    \label{fig:mode_directedness}
\end{figure}

%% file: sections/5_discussion.tex
\section{Discussion}
\label{sec:discussion}

Here, we discuss how insights into loss landscape geometry from the study of model merging can be applied to enhance model interpretability and robustness.



\subsection{Connections to Model Interpretability.} Techniques in interpretability research, such as activation steering, 
can control language style~\cite{konen2024style}, reduce model toxicity~\cite{li2024destein}, and edit factual associations post-training~\cite{meng2022locating, meng2022mass}. Such methods successfully alter model behavior via linear interventions on the activation space~\cite{wang2024adaptive, arditi2024refusallanguagemodelsmediated}, with some techniques choosing to edit weights post-training~\cite{meng2022locating, meng2022mass}. This largely activation-centric model of interpretability has yielded useful mechanistic interventions for controlling model behavior, however, the apparent brittleness of contemporary interpretability techniques under extensive evaluation makes clear the need for a novel perspective on model interpretability. Rigorous examinations have demonstrated that natural language explanations for neuron activation patterns display poor precision and recall while also failing to behave as expected under causal interventions~\cite{huang2023rigorously}. Even unsupervised interpretability techniques such as sparse autoencoders~\cite{bricken2023towards, huben2023sparse, lieberum2024gemma} have been shown to perform poorly on causal intervention benchmarks~\cite{chaudhary2024evaluating}. 
In general, activation analyses---which are dependent on inference data---may not yield robust, generalizable interpretations of model components. Insights from the study of loss landscape geometry can provide researchers with tools to circumvent these pitfalls and develop more reliable interventions to guide model behavior.

Existing model merging work on task arithmetic~\cite{ilharco2022editing}, TIES-merging~\cite{yadav2024ties}, and AdaMerge~\cite{yang2023adamerging} demonstrates the viability of weight steering as an alternative to these activation-based techniques. The established linear correspondence between changes in model activations and changes in model weights further justifies this approach~\cite{zhou2023going, zhou2024cross}. Weight steering has the additional advantage that it circumvents the need for extensive tuning, which is a critical downside to applying activation steering methods~\cite{tan2024analyzing,konen2024style,li2024destein}. For example, \citet{barbulescu2024each} demonstrate how task arithmetic can be harnessed to produce superior unlearning performance in LLMs while maintaining or even improving the underlying models' text-generation capabilities. \citet{kadhe2024split} explicitly employ model merging to augment existing unlearning techniques for LLMs by splitting their unlearning dataset, unlearning each subset individually, and merging the resulting models. This technique significantly improves unlearning capabilities while maintaining performance. 

Future work may seek to directly apply model merging techniques to studying phenomena such as memorization~\cite{carlini2023quantifying, sakarvadia2024mitigating}, overfitting~\cite{mallinar2207benign}, and catastrophic forgetting~\cite{kirkpatrick2017overcoming}. Interpretability work in these areas can draw upon insights from the model merging literature, for instance: 
\begin{enumerate*}[label=\textit{(\roman*)}]
    
    \item models trained on similar tasks or exhibiting similar behaviors, can be aligned to find common subsets of parameters for a given task \cite{ainsworth2022git, singh2020model, stoica2023zipit};
    
    \item models trained using distinct randomly chosen subsets of data or different hyperparameters, when merged, can help identify parameters in each source model that are relevant to the overall task invariant to the specific data or hyperparameters that were used to train each model \cite{wortsman2022model, zimmer2023sparse, jang2024model,  stoica2023zipit};
    
    and
    
    \item {models with similar underlying training distributions cluster together in parameter space \cite{gueta2023knowledge}}.
    
\end{enumerate*}

For example, we might choose to study how toxicity in LLMs arises by applying the methods from merging techniques such as ZipIt!~\cite{stoica2023zipit} or OTFusion~\cite{singh2020model} to identify sets of parameters across models that are responsible for toxic behavior---potentially identifying common patterns in network connectivity. Similarly, we might attempt to associate degenerate behaviors, such as memorization, with particular features of the local loss landscape geometry, developing techniques to preemptively determine whether models are at high risk for memorizing training data. Or, we might use concepts from loss landscape geometry to determine the bases over which we can best interpret our models. For instance, \citet{bushnaq2024using} explicitly invoke ideas from the study of loss landscapes to formulate a novel, more interpretable neural network parameterization. Exploring the intersection between model merging, loss landscape geometry, and interpretability is a fruitful direction for future interpretability research.

\subsection{Connections to Model Robustness}  
Model merging techniques can aid practitioners in associating model components with semantic behaviors, such as sensitivity with respect to inputs and susceptibility to adversarial examples (e.g., robustness). Many deep learning applications that must defend against \textit{adversaries}---malicious actors seeking to extract sensitive data, degrade model performance, or abuse models for inappropriate use cases---can benefit from developing novel, reliable methods to ensure model robustness. For example, in sensitive domains like healthcare, the consequences of poor model performance in diagnostic and prognostic tasks can be life-threatening---these models must be robust to adversaries~\cite{finlayson2019adversarial}. Healthcare providers may seek to accelerate medical research by using decentralized machine learning techniques, such as federated learning~\cite{li2021survey}, to create robust models without directly sharing patient data between facilities~\cite{bogdanova2020risk}. Unfortunately, decentralized model updates may risk unintentionally leaking private data by sending parameters or gradient updates across a network~\cite{dimitrov2022data} and allowing the introduction of adversarial examples that degrade overall model performance \cite{bogdanova2020risk, kuo2022detecting, darzi2023fed, zhang2019adversarial}. The model merging literature can help us explain the root causes of these vulnerabilities and provide techniques to prevent them. For example, \citet{shokri2017membership} demonstrate that adversaries can exploit mode directedness to perform membership inference attacks on black-box models hosted in the cloud, additionally showcasing the effectiveness of mitigations such as increasing prediction entropy or applying regularization in thwarting potential adversaries. Model merging techniques such as \textit{RegMean}~\cite{jin2022dataless} offer procedures for decentralized model merging that avoid the risk of leaking data samples. Furthermore, several recent studies in model robustness have established the effectiveness of model merging in improving existing models to be more robust and performant~\cite{fradkin2022robustness, zhou2023going, xu2019understanding}.

Critically, the principles of mode determinism and mode directedness are implicitly or explicitly applied in various attacks on deep learning models. When attacking black-box models, it is common to train local substitute models of a similar architecture on training data for the same underlying task to develop adversarial examples \cite{papernot2017practical, bhambri2019survey}. This assumption of similarity in learned representation and behavior is encapsulated by the concept of mode directedness and is supported by the model merging and loss landscape literature. Many attacks, such as data poisoning and membership inference, can be aided by training similar model architectures on similar data distributions~\cite{shokri2017membership}. Mode determinism implies that such attacks may be especially effective when deployed against freely available pre-trained checkpoints, as fine-tuned models stemming from the same pre-trained checkpoint will likely learn very similar representations and lie close together in the objective landscape~\cite{frankle2020linear, gueta2023knowledge}. This is especially a concern for fully open-source model checkpoints, as attackers can develop gradient-based attacks (i.e., developing adversarial examples) with some confidence that those attacks will work as intended against other fine-tuned models, even if they are fine-tuned on different datasets. For example, the membership inference attack proposed by \citet{shokri2017membership} attempts to learn whether a particular data record is present in a black-box model's training dataset by training several \enquote{shadow models} to approximate the behavior of the target model. This attack already shows great effectiveness with only black-box access to the target model; giving adversaries access to a pre-trained version of the target model would significantly increase the attack's severity. Similarly, \citet{eustratiadis2022attacking} propose a gradient-based attack that can circumvent common inference-time defenses such as gradient obfuscation by estimating a smoothened version of the local loss landscape geometry around the target model. However, such gradient-based attacks might be trivially feasible given a set of open-source model weights of the same architecture that an adversary could use to develop adversarial examples. Given the popularity of the pre-train fine-tune paradigm, particularly for LLMs, there may be significant security risks associated with deploying open-source models into production scenarios.

Overall, the study of model robustness is increasingly dependent on understanding loss landscape geometry~\cite{zhao2020bridging, xu2019understanding}.
For example, the success of gradient obfuscation against gradient-based adversaries has been linked to how gradient obfuscation alters the local loss landscape geometry to be more coarse---making it difficult to ascend for adversaries. Conversely, adversaries can effectively circumvent these defenses by navigating a smoothed surrogate loss geometry~\cite{eustratiadis2022attacking}. Train-time defenses against adversarial examples frequently involve navigating toward robust local minima by including an adversarial robustness term in the loss---one can even thwart subsequent fine-tuning success in restricted domains by appropriately engineering the loss function~\cite{deng2024sophon}. This has been observed to bias models toward large, flat loss basins. Indeed, a strong correlation has been established between adversarial robustness and local loss curvature, indicating that solutions in large, flat basins are more robust to adversaries~\cite{zhao2020bridging}. Some recent works even explicitly exploit mode connectivity to learn paths between vulnerable solutions that contain more robust models---outperforming fine-tuning approaches in terms of model robustness and accuracy~\cite{zhao2020bridging}. This underscores the advantages that a deep understanding of loss landscape geometry may yield in enhancing model robustness.



% ...

%% file: sections/6_future_work.tex
\section{Conclusion \& Future Work}
\label{sec:future_work}

We have surveyed a range of model merging works and extracted insights into the underlying structure of loss landscapes---namely by characterizing mode convexity, mode determinism, mode directedness, and mode connectivity. We discussed the intimate connections between these observations and model interpretability and robustness, offering analogs for traditional activation space analysis in weight space.

Future work exploring the intersection of these fields might focus on understanding the relationship between the objective geometry of hierarchically related tasks to understand why semantically similar tasks produce solutions close together in parameter space, even when their raw data distributions (over, say, tokens) might seem quite dissimilar \cite{gueta2023knowledge}. Such investigations would touch on the ability of neural networks to abstract and generalize across tasks in a meaningful way. One could also attempt to \enquote{orthogonalize} task-specific directions in weight space to obtain interpretable decompositions over directions in parameter space as an alternative to activation-based methods such as Distributed Alignment Search~\cite{geiger2024finding}, Sparse Autoencoders~\cite{bricken2023towards}, and the Interaction Basis~\cite{bushnaq2024using}.
Other extensions might further explore methods for continual learning, few-shot learning, or parameter-efficient fine-tuning through model merging techniques.

A concerning implication of our work is that, due to the principle of mode determinism, models fine-tuned from pre-trained checkpoints---particularly those fine-tuned on similar tasks---will have similar representations and lie close together in parameter space. Therefore, models stemming from freely available pre-trained checkpoints might be especially vulnerable to adversaries attempting to replicate model behavior \cite{shokri2017membership, papernot2017practical, bhambri2019survey}. This seems a significant risk, given that the dominant paradigm for model deployment is currently predicated on extensive pre-training and domain-specific fine-tuning. Further work exploring how to secure open-source models is paramount to ensuring a healthy open-source environment for deep learning models.

\break